\documentclass{article}

\PassOptionsToPackage{numbers, sort&compress}{natbib}
  \usepackage[preprint]{neurips_2026}

\usepackage[utf8]{inputenc} 
\usepackage[T1]{fontenc}    
\usepackage{hyperref}       
\usepackage{url}            
\usepackage{booktabs}       
\usepackage{amsfonts}       
\usepackage{nicefrac}       
\usepackage{microtype}      
\usepackage[table]{xcolor}  
\usepackage{graphicx}       
\usepackage{diagbox}        
\usepackage{amsmath}
\usepackage{dsfont}            
\usepackage{algorithm}
\usepackage{algpseudocode}
\usepackage{multirow}

\definecolor{famblue}{HTML}{12508F}
\definecolor{famamber}{HTML}{E8853A}

\title{MLIPDiff: Guiding Denoising Diffusion Models for Structure-Based Drug Design via Machine-Learned Interatomic Potentials}

\author{%
  Yiyan Liao\thanks{Equal contribution.} \\
  Peking University\\
  \texttt{yiyanliao@stu.pku.edu.cn} \\
  \And
  Amin Tavakoli\footnotemark[1] \\
  Caltech \\
  \texttt{amint@caltech.edu} \\
  \And
  Yu Zhang \\
  Caltech \\
  \texttt{yuzhang@caltech.edu} \\
  \AND
  Anima Anandkumar \\
  Caltech \\
  \texttt{anima@caltech.edu} \\
}

\begin{document}

\maketitle

\begin{abstract}
    Diffusion models have become a dominant approach to structure-based drug design. However, because they are trained primarily to fit reference structures, they often place ligands in physically implausible poses that clash with the protein. We introduce MLIPDiff, an inference-time guidance method that can be applied to pretrained diffusion models for ligand generation. The method uses physics-informed guidance from UMA (Universal Models for Atoms), a machine-learned interatomic potential (MLIP). The guidance combines four weighted energy terms (protein-ligand interaction, ligand strain, steric clash, and pocket anchoring) and uses the original model checkpoint without retraining. On CrossDocked2020, MLIPDiff completely eliminates protein-ligand collisions while preserving the ligand's local geometry, improving bond-length distributions and physical plausibility relative to the base model. Compared to models such as NucleusDiff that are specifically designed to reduce atomic clashes, MLIPDiff achieves clash-free structures without compromising local geometry. It maintains the base model's PoseBusters pass rate and attains the best median Vina Score, Vina Min, and Vina Dock, surpassing both the base model and NucleusDiff, with a median of 59.6\% of ligands per pocket binding more strongly than the reference. MLIPDiff thus addresses a limitation of prior methods by generating physically plausible structures with favorable predicted binding affinity using a single frozen model and no additional training.
\end{abstract}

\section{Introduction}

Structure-based drug design (SBDD), which generates small-molecule ligands with high binding affinity to a given protein pocket, is a fundamental task in drug discovery~\citep{wang2025modeling}. Numerous generative models have been proposed for this task~\citep{ragoza2022ligan, luo2021ar, peng2022pocket2mol, liu2022graphbp, huang2024pmdm, guan2023targetdiff}. Building on advances in probabilistic diffusion~\citep{ho2020ddpm, hoogeboom2021multinomial} and equivariant neural networks~\citep{fuchs2020se3, satorras2021egnn}, diffusion models~\citep{guan2023targetdiff, huang2024pmdm} have become the dominant paradigm. By progressively denoising from noise toward the data distribution, they achieve strong fitting capacity and refine all atoms jointly at every step. Yet they are trained only to fit that distribution. On unseen pockets they often produce sterically invalid geometries, including clashes with the protein~\citep{liu2025nucleusdiff} and distorted local geometry, undermining confidence in the predicted binding poses.

Several recent methods incorporate physical priors into diffusion-based SBDD to reduce clashes and improve physical plausibility. NucleusDiff~\citep{liu2025nucleusdiff} jointly denoises the ligand atoms and a set of mesh points that discretize atoms' van der Waals surfaces, with a regularization term restricting the atom-to-mesh distances, driving toward nearly collision-free structures. Lai et al.~\citep{lai2026forcefield} instead steer the inference process with a force field. However, both rely on classical descriptions of interatomic interactions, fixed van der Waals radii, and empirical force fields, which are crude surrogates for the true energy of a molecular system and cannot capture the finer geometric and electronic effects that govern physical plausibility. These approximations motivate guidance from a more accurate, fine-grained physical model.

Density functional theory (DFT)~\citep{kohn1965self} provides such a model, yielding the total energy and per-atom forces, so that following these forces relaxes a structure toward a physically realistic, low-energy conformation. However, DFT is far too costly to run the thousands of times a sampling loop requires. Machine-learned interatomic potentials (MLIPs)~\citep{behler2007generalized} approximate its energies and forces at near-DFT accuracy and at orders-of-magnitude lower computational cost, and foundational models such as UMA (Universal Models for Atoms)~\citep{wood2025uma} and OrbitAll~\citep{kang2025orbitall} generalize across chemical space without per-system fitting. Diffusion sampling is naturally suited to such guidance, as its reverse process predicts a clean structure at each step ($\hat{x}_0$-prediction), on which an MLIP force can be computed and used to steer the inference.

Building on these observations, we introduce MLIP-based guidance to steer pretrained diffusion models at inference time. At each denoising step, the predicted clean ligand is scored by four physically motivated energy terms, two evaluated with an MLIP (protein-ligand interaction and ligand strain) and two evaluated analytically (steric clash and pocket anchoring), whose gradients shift the reverse-process mean. We instantiate the scheme on the pretrained TargetDiff~\citep{guan2023targetdiff} backbone, yielding MLIPDiff, and evaluate it on CrossDocked2020~\citep{francoeur2020crossdocked}. Our main contributions are as follows:

\begin{itemize}
  \item \textbf{Quantum-mechanical (QM)-level physical guidance for any molecular diffusion model.} The guidance acts purely at inference on a frozen base model, requiring neither retraining nor architectural change.
  \item \textbf{Physical plausibility and affinity improve jointly.} Guidance eliminates protein-ligand collisions at no conformational cost, matching the specialized NucleusDiff while preserving the local geometry that NucleusDiff sacrifices, and it surpasses both NucleusDiff and the base model TargetDiff on predicted binding affinity, achieving a combination of physical plausibility and affinity that neither attains.
  \item \textbf{Interpretable and independently tunable objectives.} Because the guidance is an explicit sum of physical energy terms, each term has an interpretable physical role, so the objectives can be tuned independently rather than through a single score.
\end{itemize}


\section{Methods}

\subsection{Diffusion for structure-based drug design}
A protein pocket is represented as a set of atoms $\mathcal{P}=\{(x_P^{(i)}, v_P^{(i)})\}_{i=1}^{N_P}$, and the SBDD task is to generate a ligand $L=\{(x^{(i)}, v^{(i)})\}_{i=1}^{N_L}$ conditioned on it, where $x\in\mathbb{R}^3$ denotes Cartesian coordinates and $v$ denotes atom types. We build MLIPDiff on TargetDiff~\citep{guan2023targetdiff}, which diffuses the coordinates with Gaussian noise and the types with uniform noise over the $K$ categories, following a fixed variance schedule $\beta_1,\dots,\beta_T$ with $\alpha_t=1-\beta_t$ and $\bar\alpha_t=\prod_{s=1}^{t}\alpha_s$, so that
\begin{equation}
q(x_t \mid x_0) = \mathcal{N}\!\big(x_t;\, \sqrt{\bar\alpha_t}\, x_0,\, (1-\bar\alpha_t) I\big).
\label{eq:forward}
\end{equation}
Its SE(3)-equivariant network predicts the clean ligand directly,
$[\hat{x}_0, \hat{v}_0] = \phi_\theta([x_t, v_t], t, \mathcal{P})$, and each reverse step samples
$x_{t-1}\sim\mathcal{N}(\mu_\theta(x_t,\hat{x}_0), \sigma_t^2 I)$ from the Gaussian posterior of
equation~\ref{eq:forward}, with the types drawn from the corresponding categorical posterior
$\tilde{c}_t(v_t,\hat{v}_0)$. The one-step denoised prediction $[\hat{x}_0, \hat{v}_0]$ available
at every step is the quantity our guidance acts on.

\begin{figure}[!t]
    \centering
    \includegraphics[width=1\linewidth]{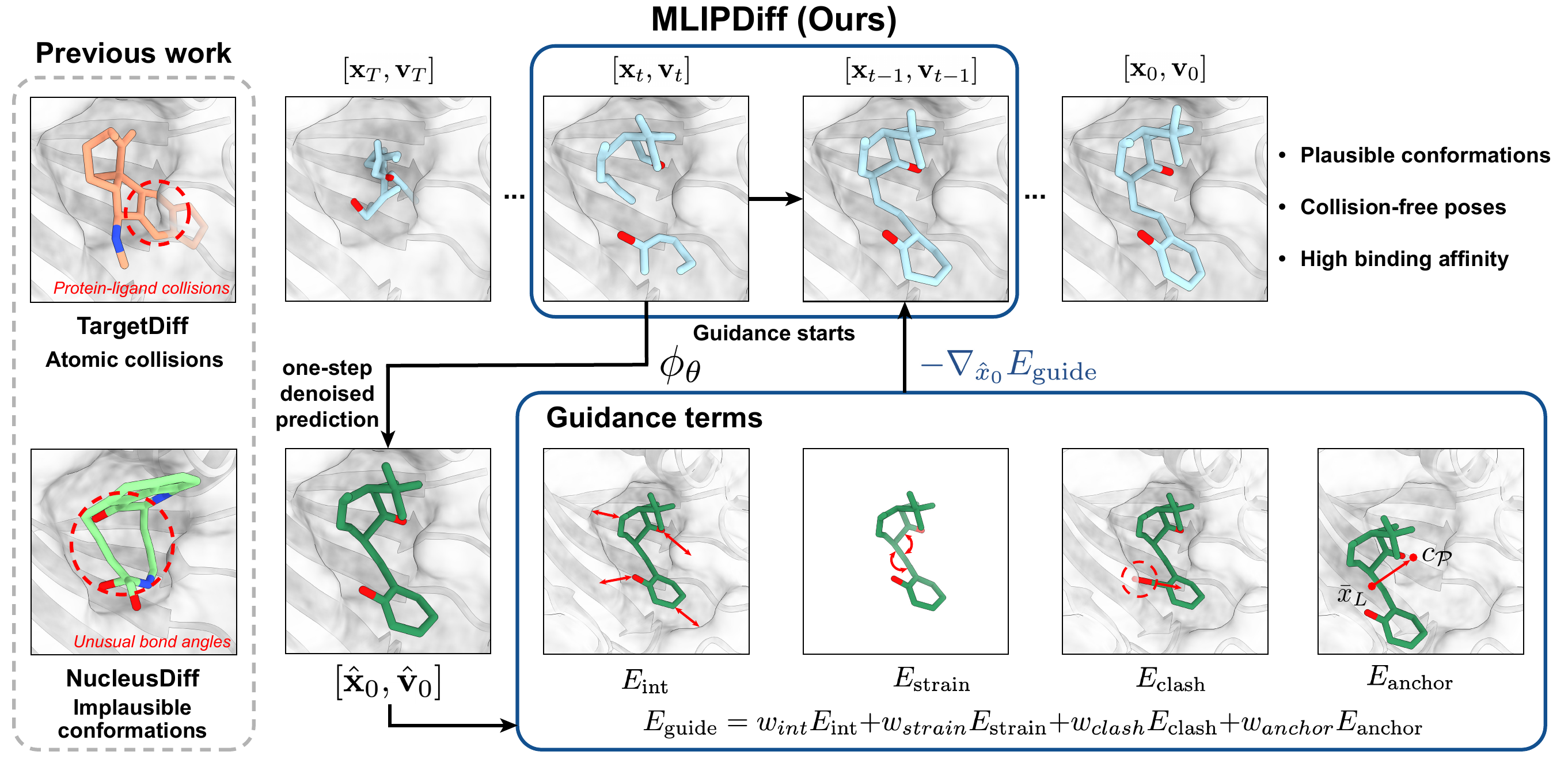}
    \caption{Overview of MLIPDiff. A sampling trajectory of MLIPDiff (light blue), with guidance turned on when $t$ falls below a threshold. At each step $\phi_\theta$ predicts the one-step denoised ligand $[\hat{x}_0, \hat{v}_0]$ (sea green), on which four guidance terms yield a gradient that shifts the mean of the reverse-process posterior. TargetDiff (light salmon), the same backbone without guidance, ends in atomic collisions with the pocket (red circles). NucleusDiff (light green) produces an implausible conformation (red circle).}
\label{fig:figure1}
\end{figure}

\subsection{Guiding diffusion models with MLIP}

Classifier guidance~\citep{dhariwal2021diffusion} steers a pretrained diffusion model toward a desired property $y$ by guiding each reverse step with $\nabla_{x_t}\log p(y\mid x_t)$. We take physical plausibility as the property and express it through an energy, setting $\log p(y\mid x_t)\propto -E_\text{guide}$, so that the gradient descends the energy surface. We calculate this energy with UMA~\citep{wood2025uma}, a universal MLIP that reproduces DFT energies and forces across chemical space.

However, an MLIP is meaningful only for chemically valid geometries, which the noised state $x_t$ lacks, especially at high $t$ (early steps). The one-step denoised prediction $\hat{x}_0$, by contrast, is exactly the object the potential was designed for. We therefore evaluate $E_\text{guide}$ on $[\hat{x}_0, \hat{v}_0]$~\citep{chung2023dps}, and approximate $\nabla_{x_t}E_\text{guide}$ by $\nabla_{\hat{x}_0}E_\text{guide}$. Moreover, the prediction becomes reliable only after sufficient noise has been removed, so guidance is activated at or below a threshold $t_\text{thr}$ and the early steps are left to the diffusion prior alone. We choose $t_\text{thr}=400$ in our implementation, justified in Appendix~\ref{si:gating}.

Two further considerations determine the guidance step size. An MLIP force is expressed in eV/\AA{} while the sampler moves in units set by the noise schedule, so the raw gradient must be rescaled before being added to a diffusion step; and a guidance term that dominates the update would drag samples off the manifold the model has learned, producing chemically implausible structures. We address the first by multiplying the gradient by the step's own noise scale $\sigma_t=\sqrt{\tilde\beta_t}$ and a cosine ramp
$k(t)=\cos^2\!\big(\frac{t}{T}\frac{\pi}{2}\big)$, which vanishes at $t=T$ and reaches one at
$t=0$, so that guidance is suppressed at high noise and brought to full strength as the structure
takes shape:
\begin{equation}
g_t = k(t)\, \sigma_t\, \nabla_{\hat{x}_0} E_\text{guide}.
\label{eq:gt}
\end{equation}
The factor $k(t)\,\sigma_t$ is applied uniformly to every term; the per-term strengths are set by the weights $w$ in Eq.~\ref{eq:energy}. We address the second by capping the guidance against the model's own displacement, never allowing the correction to exceed $\lVert\mu_\theta - x_t\rVert\,k(t)$, the distance the diffusion process itself travels at that step. Writing $c_t=\min\{1,\ \lVert\mu_\theta - x_t\rVert\,k(t)/(\lVert g_t\rVert+\delta)\}$ for the resulting factor, with $\delta$ a small constant, the guided step shifts the posterior mean $\mu_\theta$,
\begin{equation}
x_{t-1} \sim \mathcal{N}\!\big(\mu_\theta - c_t g_t,\ \sigma_t^2 I\big) .
\label{eq:guided}
\end{equation}
The cap means guidance can steer a trajectory but never override the learned prior, and because the correction enters only the mean, the variance is untouched and the sampling variance remains unchanged. The atom types are left to the unmodified categorical posterior. The whole scheme runs on a frozen checkpoint at no training cost. Figure~\ref{fig:figure1} gives an overview, and Algorithm~\ref{alg:sampling} summarizes the procedure.

\begin{algorithm}[!h]
\caption{MLIPDiff sampling}
\label{alg:sampling}
\begin{algorithmic}[1]
\Require Pocket $\mathcal{P}$, network $\phi_\theta$, MLIP $U$, weights $w_\text{int},w_\text{strain},w_\text{clash},w_\text{anchor}$, onset $t_\text{thr}$
\State Centre $\mathcal{P}$ at the origin; sample $N_L$ from the size prior
\State $x_T \sim \mathcal{N}(0, I)$; \ $v_T \sim$ uniform over the $K$ atom types
\For{$t = T, \dots, 1$}
  \State $[\hat{x}_0, \hat{v}_0] \gets \phi_\theta([x_t, v_t], t, \mathcal{P})$
    \Comment{one-step denoised prediction}
  \State $\mu_\theta \gets \mu_\theta(x_t, \hat{x}_0)$; \ $\sigma_t \gets \sqrt{\tilde\beta_t}$
    \Comment{DDPM posterior}
  \If{$t \le t_\text{thr}$}
    \State Protonate $[\hat{x}_0, \hat{v}_0]$; compute $\nabla_{\hat{x}_0} E_\text{guide}$ with the MLIP $U$
    \State $g_t \gets k(t)\,\sigma_t\,\nabla_{\hat{x}_0} E_\text{guide}$; \
      $c_t \gets \min\!\big\{1,\ \lVert\mu_\theta-x_t\rVert\,k(t)/(\lVert g_t\rVert+\delta)\big\}$
    \State $\mu_\theta \gets \mu_\theta - c_t g_t$
      \Comment{guidance shift}
  \EndIf
  \State $x_{t-1} \sim \mathcal{N}(\mu_\theta, \sigma_t^2 I)$; \
    $v_{t-1} \sim \mathcal{C}(\tilde{c}_t(v_t, \hat{v}_0))$
\EndFor
\State \Return molecule reconstructed from $[x_0, v_0]$
\end{algorithmic}
\end{algorithm}

\subsection{A decomposed guidance energy}
\label{sec:terms}
A single energy term could conflate distinct failure modes. A generated ligand can bind weakly, strain its own bonds, or overlap the protein, and these call for different corrections. We therefore decompose $E_\text{guide}$ into four physically motivated terms, each targeting a different aspect of molecular quality:
\begin{equation}
E_\text{guide} = w_\text{int} E_\text{int}
+ w_\text{strain}
E_\text{strain} + w_\text{clash}
E_\text{clash} + w_\text{anchor}
E_\text{anchor}
\label{eq:energy}
\end{equation}
The interaction term $E_\text{int}=E_\text{MLIP}(\mathcal{P}\cup L)-E_\text{MLIP}(\mathcal{P})-E_\text{MLIP}(L)$ is the binding energy and rewards favorable protein-ligand interactions. The target-aware prior already positions the ligand within the pocket, but from learned pose statistics rather than an explicit energy, so $E_\text{int}$ supplies the physical binding signal that drives the ligand toward configurations with stronger predicted affinity. The strain term $E_\text{strain}=E_\text{MLIP}(L)$ is the internal energy of the isolated ligand. Because a favorable complex energy can be attained by trading strained bonds and torsions for improved contacts, we also score the isolated ligand, penalizing internal distortion used to compensate for weak interactions. The clash term $\sum_{i\in L}\sum_{j\in\mathcal{P}}[\max(0, r^{\min}_{ij}-r_{ij})]^2$ is activated only when a protein-ligand distance $r_{ij}$ falls below the sum $r^{\min}_{ij}$ of the two van der Waals radii. We keep it as a separate analytic penalty rather than relying on the MLIP to resolve overlaps, because the MLIP is least reliable exactly where atoms overlap, a configuration it rarely saw during training. The anchor term $\frac{1}{2}\lVert\bar{x}_L-c_\mathcal{P}\rVert^2$ restrains the ligand centroid $\bar{x}_L$ toward the pocket center $c_\mathcal{P}$, a defensive term that keeps the ligand from drifting away from the pocket while the other terms act.

The two MLIP terms are computed with UMA through an atomistic-simulation interface that returns forces without exposing the autograd graph (see Appendix~\ref{si:uma} for the checkpoint and inference settings). The force returned by UMA is the negative gradient of its predicted energy, $F=-\nabla_{x}E_\text{UMA}$, computed by autograd through the potential, so we negate the returned forces to obtain $\nabla_{\hat{x}_0}E$ for the two MLIP terms. The clash and anchor terms are analytic and differentiated by autograd. The weights $w$ balance the four contributions; we report the calibration procedure and the values used in Appendix~\ref{si:calibration}.

\section{Experiments}


\subsection{Dataset}

We use the CrossDocked2020~\citep{francoeur2020crossdocked} test set as our benchmark to assess the quality of ligands generated for the target proteins. Following prior work~\citep{guan2023targetdiff, zhang2026denovo}, we use its 100 proteins for testing and sample 100 molecules per pocket, with the number of atoms drawn from a prior distribution conditioned on pocket size.

\subsection{Metrics}

We consider several categories of metrics to evaluate the generated molecules, including conformational plausibility, atomic collisions, binding affinity, and drug-likeness.

\textbf{Conformational plausibility.} We report the Jensen--Shannon divergence (JSD) between the bond-distance distributions of generated and reference ligands, computed per bond type (lower is closer to the reference geometry); the median root-mean-square deviation (RMSD) of rigid fragments (those with no rotatable bonds) before and after force-field optimization, which measures how close the local geometry is to a relaxed conformation; and the PoseBusters pass rate (PBR), the fraction of ligands that pass all PoseBusters checks~\citep{buttenschoen2024posebusters} of parsing, bond lengths and angles, internal clashes, and ring planarity, where a single failure fails the molecule.

\textbf{Atomic collisions.} The pairwise-level collision ratio (PLCR), atom-level collision ratio (ALCR), and molecule-level collision ratio (MLCR) count ligand-pocket atom collisions at the pairwise, atom, and molecule level, where two atoms clash when their distance is below the sum of their covalent radii, a more strict quantity involving the covalent bonding, following NucleusDiff~\citep{liu2025nucleusdiff}; we track them across sampling steps (see Appendix~\ref{si:collision} for definitions). 

\textbf{Binding affinity and drug-likeness.} We use the standard AutoDock Vina~\citep{trott2010autodock} energies, namely Vina Score on the generated pose, Vina Min after local minimization, and Vina Dock after redocking, where high affinity denotes the fraction of generated molecules that outperform the reference according to Vina Dock. We further report the quantitative estimate of drug-likeness (QED)~\citep{bickerton2012quantifying}, synthetic accessibility (SA)~\citep{ertl2009sa, you2018gcpn}, and diversity, defined as the mean pairwise Tanimoto distance among a pocket's molecules~\citep{tanimoto1958, bajusz2015tanimoto}.

\subsection{Baselines}

For benchmarking, we compare with liGAN~\citep{ragoza2022ligan}, GraphBP~\citep{liu2022graphbp}, AR~\citep{luo2021ar}, and Pocket2Mol~\citep{peng2022pocket2mol}, together with the diffusion models TargetDiff~\citep{guan2023targetdiff} and NucleusDiff~\citep{liu2025nucleusdiff}. Our MLIPDiff uses the same checkpoint as TargetDiff, and NucleusDiff uses the same underlying diffusion architecture. All generated molecules pass through the same evaluation pipeline for a fair comparison.

\subsection{Results}

\subsubsection{Conformational plausibility}

Physical guidance should not perturb the local geometry the base model has learned. A well-calibrated model should reproduce the reference structural distributions, so its molecules obey dataset properties like bond-length distributions and have rigid fragments that are stable under force-field optimization. We therefore tested the bond-distance JSD relative to reference ligands and the median RMSD of rigid fragments before and after force-field optimization.

Across most major bond types, MLIPDiff improves over its base model TargetDiff (Table~\ref{tab:JSD}). NucleusDiff, by contrast, performs markedly worse, indicating that its restriction strategy shifts the bond-length distributions away from those of the reference ligands and compromises local geometry.

\begin{table}[!h]
    \centering
    \renewcommand{\arraystretch}{1.1} 
    {\small
    \caption{Jensen-Shannon divergence between the bond-length distributions of reference and generated molecules ($\downarrow$ better). ``$-$'', ``$=$'', and ``:'' represent single, double, and aromatic bonds, respectively.}
    \label{tab:JSD}
    \begin{tabular}{c|cccccc}
        \toprule
        Bond & liGAN & AR & Pocket2Mol & TargetDiff & NucleusDiff & MLIPDiff (Ours)\\ 
        \midrule
        C$-$C & 0.601 & 0.609 & 0.496 & 0.295 & 0.500 & \textbf{0.280}\\
        C$=$C & 0.665 & 0.620 & 0.561 & 0.190 & 0.426 & \textbf{0.186}\\
        C$-$N & 0.634 & 0.474 & 0.416 & 0.238 & 0.395 & \textbf{0.230}\\
        C$=$N & 0.749 & 0.635 & 0.629 & \textbf{0.163} & 0.460 & 0.170\\
        C$-$O & 0.656 & 0.492 & 0.454 & 0.297 & 0.458 & \textbf{0.290}\\
        C$=$O & 0.661 & 0.558 & 0.516 & 0.397 & 0.417 & \textbf{0.383}\\
        C:C & 0.497 & 0.451 & 0.416 & 0.199 & 0.515 & \textbf{0.197}\\
        C:N & 0.638 & 0.552 & 0.487 & 0.121 & 0.388 & \textbf{0.117}\\
        \bottomrule
    \end{tabular}}
\end{table}

\begin{figure}[!h]
    \centering
    \includegraphics[width=0.9\textwidth]{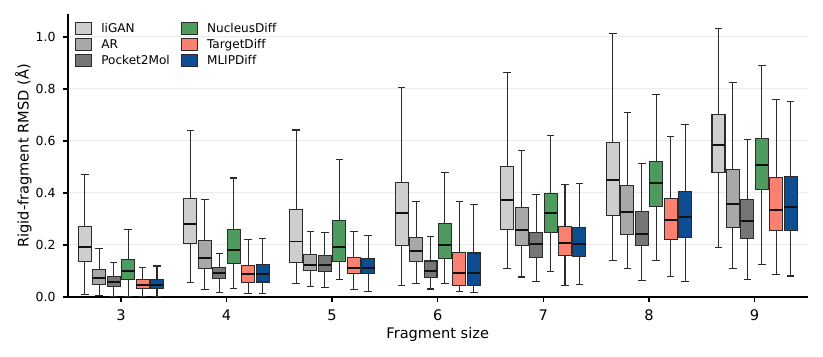}
    \caption{Median RMSD of rigid fragments before and after force-field optimization, by fragment size (number of heavy atoms).}
    \label{fig:rigid_frag}
\end{figure}

The rigid-fragment RMSD tells the same story (Figure~\ref{fig:rigid_frag}). A rigid fragment is a maximal substructure with no rotatable bonds, such as a ring or conjugated group, obtained by cutting the molecule at each rotatable bond. Optimizing the generated structures with a force field and measuring the RMSD between pre- and post-optimization coordinates of these fragments, we find that MLIPDiff closely matches TargetDiff, whereas NucleusDiff is roughly twice as high at every fragment size.

\subsubsection{Protein-ligand collisions}

MLIP guidance drives protein-ligand collisions to zero, a common form of physical implausibility. To isolate this effect from architectural differences, we compare only with TargetDiff and NucleusDiff.

\begin{figure}[h!]
    \centering
    \includegraphics[width=\textwidth]{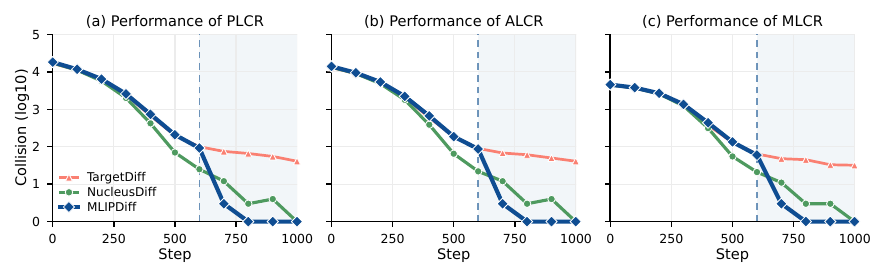}
    \caption{Protein-ligand collision counts across sampling steps, at the atom-pair, atom,
    and molecule level (log scale). The blue dashed line marks the guidance onset (step 600,
    $t\le t_{\text{thr}}=400$).}
    \label{fig:collision}
\end{figure}

Over the first 600 steps, MLIPDiff has essentially the same collision ratios as TargetDiff (Figure~\ref{fig:collision}). Once guidance is activated, collision ratios drop rapidly and reach zero by sampling step 800. NucleusDiff also reaches collision-free results, as reported. Notably, we obtain the same outcome directly with a more realistic physical prior, without the mesh-point diffusion and retraining that NucleusDiff requires.

Read together with the conformational results, NucleusDiff's collision-free structures come at the cost of a collapse in local geometry, whereas MLIPDiff eliminates the same clashes while preserving the base model's local geometry.

\subsubsection{Binding affinity and drug-likeness}

On physically plausible geometries, the guidance also improves binding affinity. We reasoned that physical information should help diffusion models capture protein-ligand interactions and generate higher-potency molecules, so we next evaluated affinity and drug-likeness. MLIPDiff improves over its base model TargetDiff on every Vina metric and attains the best median Vina Score, Vina Min, and Vina Dock among all baselines (Table~\ref{tab:metrics_comparison}); it also achieves the highest median high-affinity fraction (59.6\%), indicating more favorable predicted docking scores for the generated molecules. 

\begin{table*}[!htb]
    \centering
    \renewcommand{\arraystretch}{1.3}
    \caption{Binding affinity and molecular properties on CrossDocked2020 (100 pockets, 100 ligands each). Vina energies in kcal/mol ($\downarrow$ better); best in bold, second best underlined. For liGAN and GraphBP ($*$), AutoDock Vina could not parse some generated atom types and thus we do not provide their Vina Score and Vina Min results.}
    \label{tab:metrics_comparison}
    \vspace{1.5ex}
    \resizebox{\textwidth}{!}{%
    \begin{tabular}{lccccccccccccccc}
        \toprule
        \diagbox{Model}{Metric} & \multirow{2}{*}{PBR} & \multicolumn{2}{c}{Vina Score ($\downarrow$)} & \multicolumn{2}{c}{Vina Min ($\downarrow$)} & \multicolumn{2}{c}{Vina Dock ($\downarrow$)} & \multicolumn{2}{c}{High Affinity ($\uparrow$)} & \multicolumn{2}{c}{QED ($\uparrow$)} & \multicolumn{2}{c}{SA ($\uparrow$)} & \multicolumn{2}{c}{Diversity ($\uparrow$)} \\
        \cmidrule(lr){3-4} \cmidrule(lr){5-6} \cmidrule(lr){7-8} \cmidrule(lr){9-10} \cmidrule(lr){11-12} \cmidrule(lr){13-14} \cmidrule(lr){15-16}
         & & Avg. & Med. & Avg. & Med. & Avg. & Med. & Avg. & Med. & Avg. & Med. & Avg. & Med. & Avg. & Med. \\
        \midrule
        liGAN *           & 0.04 & -                 & -                 & -                 & -                 & -6.33             & -6.20             & 21.1\%             & 11.1\%             & 0.39             & 0.39             & 0.59             & 0.57             & 0.66             & 0.67 \\
        GraphBP *         & 0.03 & -                 & -                 & -                 & -                 & -4.80             & -4.70             & 14.2\%             & 6.7\%              & 0.43             & 0.45             & 0.49             & 0.48             & \textbf{0.79}    & \textbf{0.78} \\
        AR                & \underline{0.67} & -5.75             & -5.64             & -6.18             & -5.88             & -6.75             & -6.62             & 37.9\%             & 31.0\%             & \underline{0.51}             & \underline{0.50}             & \underline{0.64}             & \underline{0.63}             & 0.70             & 0.70 \\
        Pocket2Mol        & \textbf{0.88} & -5.14             & -4.70             & -6.42             & -5.82             & -7.15             & -6.79             & 48.4\%             & 51.0\%             & \textbf{0.57} & \textbf{0.58} & \textbf{0.76}    & \textbf{0.76}    & 0.69             & 0.71 \\
        NucleusDiff       & 0.12 & \underline{-6.11}    & \underline{-6.74}    & \underline{-6.96}    & \underline{-6.81} & \textbf{-7.88}    & \underline{-7.77}             & \underline{57.3\%} & \underline{56.2\%}             & 0.40             & 0.40             & 0.53             & 0.52             & 0.74             & 0.73 \\
        TargetDiff        & 0.62 & -5.38             & -6.03             & -6.51             & -6.57             & -7.56             & -7.65             & 53.4\%             & 52.8\%             & 0.49             & 0.49             & 0.60             & 0.59             & \underline{0.75}             & \underline{0.74} \\
        MLIPDiff (Ours) & 0.62 & \textbf{-6.13} & \textbf{-6.75} & \textbf{-6.97} & \textbf{-7.02}    & \underline{-7.64}             & \textbf{-7.87} & \textbf{58.5\%}    & \textbf{59.6\%}    & 0.48             & 0.48             & 0.58             & 0.58             & 0.72             & 0.71 \\
        \midrule
        Reference         & 0.92 & -6.36             & -6.46             & -6.71             & -6.49             & -7.45             & -7.26             & -                  & -                  & 0.48             & 0.47             & 0.73             & 0.74             & -                & - \\
        \bottomrule
    \end{tabular}%
    }
\end{table*}

The same table reports the PBR, which measures whether a pose is geometrically plausible, and here MLIPDiff matches TargetDiff at 0.62, whereas NucleusDiff falls to 0.12. The gap follows from how NucleusDiff avoids collisions. It surrounds each ligand atom with several mesh points at its van der Waals radius and trains the atoms to sit at that separation~\citep{liu2025nucleusdiff}, the same steric spacing AutoDock Vina rewards, which can yield favorable Vina scores. A van der Waals radius, however, only fixes how far apart two non-bonded atoms should sit, whereas bonded atoms lie much closer, so enforcing it leaves the covalent geometry unconstrained. Because Vina docking scores do not fully capture internal molecular validity, such distortions may not be adequately penalized.

Guidance leaves the drug-likeness metrics essentially unchanged, with median QED and SA values of 0.48 and 0.58, respectively, compared with 0.49 and 0.59 for the base model, as expected since QED and SA are two-dimensional graph properties that our guidance does not act on. This gap is shared by the entire diffusion family, which trails autoregressive models such as Pocket2Mol (0.58 and 0.76). NucleusDiff, by contrast, degrades both metrics to 0.40 and 0.52, departing from the distribution of drug-like molecules, whereas our guidance preserves the base model's drug-likeness.

\subsubsection{Case studies}

\begin{figure}[!bth]
    \centering
    \includegraphics[width=1\linewidth]{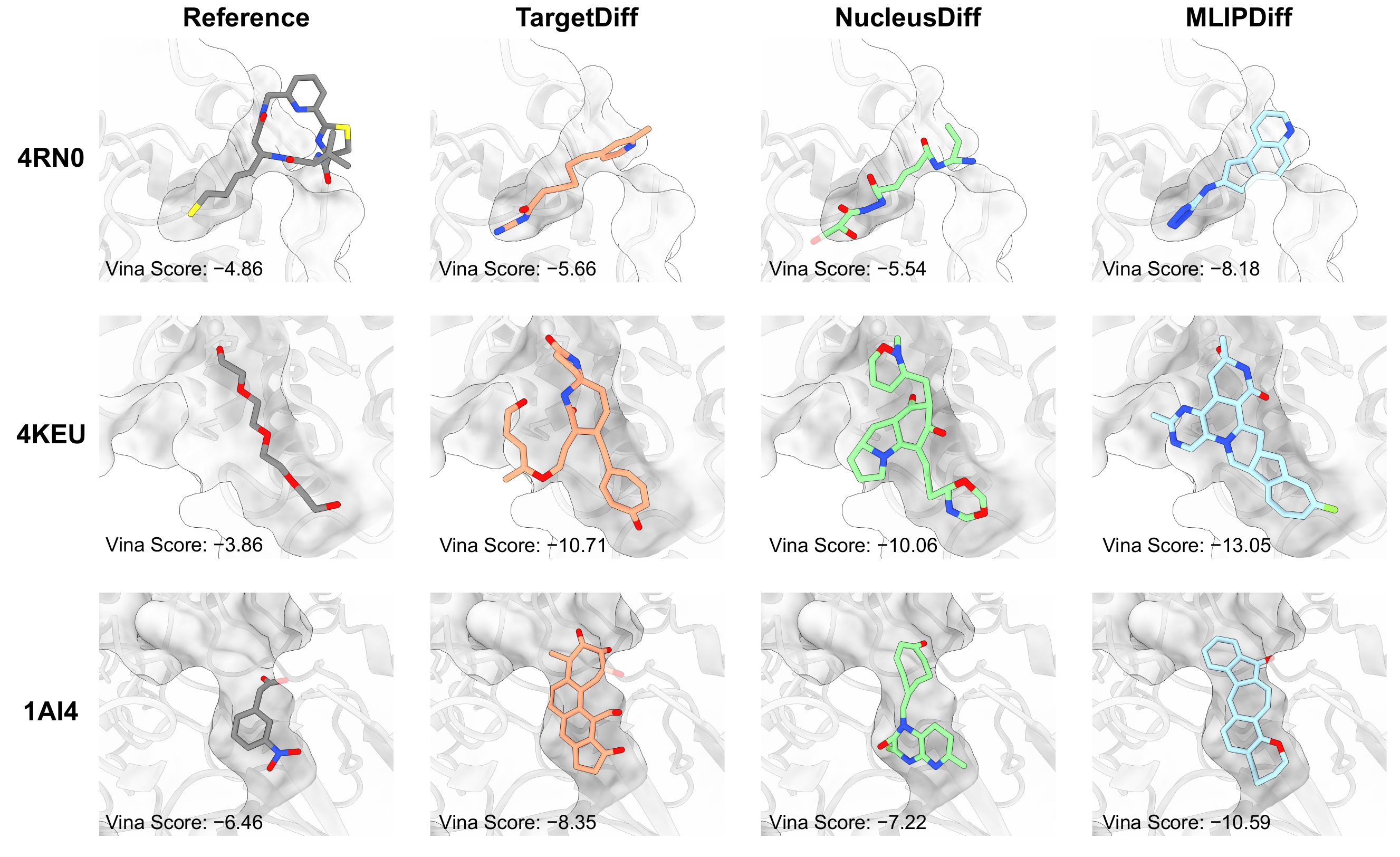}
    \caption{Visualization of the pockets and sampled ligands on the CrossDocked2020 test set. Vina scores are provided to indicate binding affinities.}
    \label{fig:case_study}
\end{figure}

We visualize three representative pockets (PDB 4RN0, 4KEU, and 1AI4; Figure~\ref{fig:case_study}), each showing a molecule from TargetDiff, NucleusDiff, and MLIPDiff matched in heavy-atom count across the three methods, with the pocket's reference ligand for context. MLIPDiff attains the best Vina Score in all three, -8.18, -13.05, and -10.59 against -5.66, -10.71, and -8.35 for TargetDiff and -5.54, -10.06, and -7.22 for NucleusDiff, while the NucleusDiff molecule fails PoseBusters in every case. In 4RN0, the baselines reach the buried subpocket with only single atoms or groups, whereas MLIPDiff fills it with a tetrazole and extends a fused-ring system toward the pocket mouth. In 4KEU, the reference is a short polyethylene glycol-like oligomer that barely contacts the pocket, while MLIPDiff fills the cavity with a large rigid fused-ring system. In 1AI4, MLIPDiff again occupies the elongated pocket with a fused-ring scaffold.

\section{Conclusion}

We presented MLIPDiff, a scheme that steers a pretrained target-aware diffusion model with a universal MLIP, decomposed into four physically motivated energy terms and applied entirely at inference. The guidance brings lightweight, QM-level physical information to molecular diffusion models without retraining the backbone. It drives protein-ligand collisions to zero at no conformational cost, and at the same time achieves the best median scores among the baselines across the three Vina metrics. Because the guidance is an explicit sum of physical energy terms, each term has an interpretable physical role, giving an interpretable framework whose objectives can be tuned independently rather than through a single score.

The main limitation is conformational. While MLIPDiff removes clashes and improves affinity, its local geometry stays close to that of the base model rather than surpassing it, as reflected in a PBR and rigid-fragment RMSD that match TargetDiff but still fall short of the reference structures. Pushing conformational quality beyond the base model will require a more careful selection of the term weights and their scheduling across the denoising trajectory. Looking further ahead, because the guidance acts only on an intermediate clean prediction rather than on any particular backbone, the same approach can extend beyond a single model. Applying it to other diffusion models, and adapting it to flow-matching and autoregressive generators, is a natural route toward making physically grounded generation a general capability in structure-based drug design.

{
\small
\bibliographystyle{unsrtnat}
\bibliography{main}
}








\appendix

\section{Methods and implementation details}

\subsection{Choice of the guidance onset \texorpdfstring{$t_\text{thr}$}{t\_thr}}
\label{si:gating}

An MLIP is informative only for chemically valid geometries, so we asked how physically valid the one-step prediction $\hat{x}_0$ remains as noise increases. We randomly sampled 100 training-set ligands, noised each to level $t$, denoised it back to $\hat{x}_0$ with the atom types fixed at their reference values so that only geometry varied, protonated it, and scored it with UMA. Table~\ref{tab:gating} reports the median heavy-atom RMS force at each level, together with $n$, the number of ligands that OpenBabel reconstructs into a valid molecule and that therefore enter the median. At high $t$ the force exceeds 50~eV/\AA{} and many ligands fail to reconstruct. As $t$ falls, the force decreases toward the reference value of 3.6, coming within about 1~eV/\AA{} of it by $t=400$, the level at which reconstruction also becomes fully reliable ($n=100$). We therefore switch guidance on at $t_\text{thr}=400$, the earliest level at which $\hat{x}_0$ is reliably reconstructed and its force has approached the reference value.

\begin{table}[h]
\centering
\caption{UMA heavy-atom RMS force (eV/\AA) on $\hat{x}_0$, median over 100 randomly drawn training-set ligands at each noise level $t$. $n$ is the number of those ligands that OpenBabel reconstructs into a valid molecule.}
\label{tab:gating}
\small
\begin{tabular}{rccc c}
\toprule
$t$ & $1-\bar\alpha_t$ & Force (with H) & Force (without H) & $n$ \\
\midrule
999 & 0.632 & 106.0 & 101.2 & \phantom{0}68 \\
900 & 0.552 & 92.3 & 104.3 & \phantom{0}74 \\
800 & 0.455 & 65.4 & 98.0 & \phantom{0}78 \\
700 & 0.340 & 53.4 & 73.5 & \phantom{0}83 \\
600 & 0.217 & 33.4 & 33.3 & \phantom{0}99 \\
500 & 0.110 & 11.9 & \phantom{0}8.8 & \phantom{0}99 \\
450 & 0.071 & \phantom{0}5.7 & \phantom{0}6.1 & 100 \\
400 & 0.043 & \phantom{0}4.7 & \phantom{0}5.7 & 100 \\
300 & 0.014 & \phantom{0}3.9 & \phantom{0}5.6 & 100 \\
200 & 0.004 & \phantom{0}3.7 & \phantom{0}5.5 & 100 \\
100 & 0.001 & \phantom{0}3.5 & \phantom{0}5.5 & 100 \\
\phantom{00}0 & 0.000 & \phantom{0}3.4 & \phantom{0}5.5 & 100 \\
\midrule
\multicolumn{2}{c}{Ground-truth reference} & 3.6 & 5.6 & 100 \\
\bottomrule
\end{tabular}
\end{table}

\subsection{UMA inference details}
\label{si:uma}

All guidance forces come from the UMA-small checkpoint (\texttt{uma-s-1p2}) with its OMol (Open Molecules)~\citep{levine2025omol25} organic-molecule head. The force this checkpoint returns is the negative gradient of its predicted energy, computed by autograd through the potential rather than by an independent force head, so the guidance force is exactly $-\nabla E$. Each ligand or protein-ligand complex is scored as an isolated non-periodic system in a cubic cell sized to its spatial extent plus 15~\AA{} of vacuum. The isolated ligand is assigned a total charge of 0, and each protein-ligand complex the net charge of its protonated pocket; all systems use a spin multiplicity of 1. The ligand protonation state is unresolved during sampling, so we keep the ligand neutral throughout. UMA is an all-atom potential whereas the diffusion model carries only heavy atoms, so at each guided step we reconstruct the predicted ligand, add hydrogens at inferred positions, score the protonated molecule, and retain the forces on the heavy atoms alone, with hydrogens serving only to complete valences. The strain term is one UMA evaluation of the isolated ligand, and the interaction term subtracts the pocket and ligand energies from the complex.

\subsection{Guidance weight calibration}
\label{si:calibration}

The four terms live on incomparable scales. $E_\text{int}$ and $E_\text{strain}$ are UMA energies in eV, whereas $E_\text{clash}$ and $E_\text{anchor}$ are analytic penalties in \AA{}, so we fix each weight by a common rule. On 32 training-set complexes at $t=100$, we measure the per-atom diffusion step $\lVert\mu_\theta(x_t,\hat{x}_0)-x_t\rVert$ and the displacement each term adds at unit weight, $k(t)\,\sigma_t\lVert\nabla_{\hat{x}_0}E_k\rVert$, then set $w_k$ so the latter is a fixed fraction of the former, namely 20\% for interaction, strain, and anchor, and 40\% for clash, the term that must most strongly push atoms apart. Each contribution stays below the diffusion step, so the clip $c_t$ rarely binds (Appendix~\ref{si:guidance_strength} tracks each term's applied move across the guided window). We calibrate at low $t$ because the diffusion gradient grows by an order of magnitude toward high $t$.

\begin{table}[h]
\centering
\caption{Guidance weights used in all experiments, each calibrated to a fixed fraction of the per-atom diffusion step $\lVert\mu_\theta-x_t\rVert$ at $t=100$.}
\label{tab:calibration}
\small
\begin{tabular}{lcc}
\toprule
Term & Weight & Fraction \\
\midrule
interaction $w_\text{int}$   & $0.10$  & $20\%$ \\
strain $w_\text{strain}$     & $0.016$ & $20\%$ \\
clash $w_\text{clash}$       & $1.0$   & $40\%$ \\
anchor $w_\text{anchor}$     & $0.40$  & $20\%$ \\
\bottomrule
\end{tabular}
\end{table}

\subsection{Collision metric definitions}
\label{si:collision}

The pairwise-, atom-, and molecule-level collision ratios (PLCR, ALCR, MLCR) measure how often generated ligand atoms overlap pocket atoms, following NucleusDiff~\citep{liu2025nucleusdiff}. For each ligand atom we take its $K=10$ nearest protein atoms in the pocket, and count a ligand-protein pair $(i,j)$ as a collision when their distance falls below $D_{ij}$, the sum of their covalent radii. Writing $M$ for the number of generated molecules, $N_k$ for the atom count of molecule $k$, and $\mathcal{N}_i$ for the $K$ nearest protein atoms of ligand atom $i$, the three ratios are
\begin{align}
\mathrm{PLCR}&=\frac{\sum_{k=1}^{M}\sum_{i\in k}\sum_{j\in\mathcal{N}_i}\mathds{1}\!\left[\lVert x_i-x_j\rVert<D_{ij}\right]}{K\sum_{k=1}^{M}N_k},\\
\mathrm{ALCR}&=\frac{\sum_{k=1}^{M}\sum_{i\in k}\mathds{1}\!\left[\sum_{j\in\mathcal{N}_i}\mathds{1}\!\left[\lVert x_i-x_j\rVert<D_{ij}\right]>0\right]}{\sum_{k=1}^{M}N_k},\\
\mathrm{MLCR}&=\frac{\sum_{k=1}^{M}\mathds{1}\!\left[\sum_{i\in k}\sum_{j\in\mathcal{N}_i}\mathds{1}\!\left[\lVert x_i-x_j\rVert<D_{ij}\right]>0\right]}{M}.
\end{align}
PLCR is the fraction of protein-ligand collision pairs among all $K\sum_k N_k$ pairs, ALCR is the fraction of ligand atoms that clash with at least one nearby protein atom, and MLCR is the fraction of molecules containing at least one colliding atom.

\section{Additional results}

\subsection{Per-term guidance strength across sampling}
\label{si:guidance_strength}

\begin{figure}[!h]
    \centering
    \includegraphics[width=\linewidth]{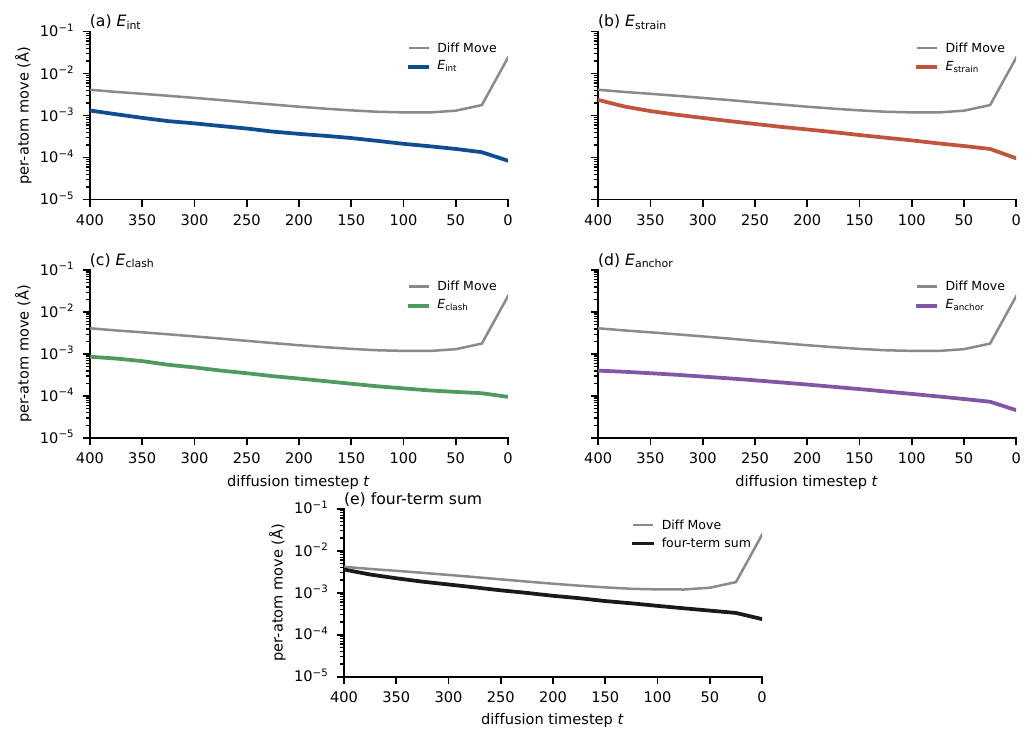}
    \caption{Per-atom guidance displacement across sampling. Each panel plots the applied per-atom move against the diffusion step size (Diff Move); panels (a) to (d) are the four energy terms and (e) their vector sum.}
    \label{fig:guidance_strength}
\end{figure}

On the sampled trajectories we compute the per-atom displacement each energy term contributes at every step, averaged over ligand atoms, alongside their vector sum and the diffusion step size $\lVert\mu_\theta-x_t\rVert$ (Diff Move). Figure~\ref{fig:guidance_strength} reports the median of these quantities over the sampled trajectories within the guided window $t\le t_\text{thr}=400$.

\end{document}